\documentclass[sigconf]{acmart}
\usepackage{tikz}
\usepackage{algorithm}
\usepackage{algpseudocode}
\usepackage{pgfplots}
\usepackage{float}   % for [H] tables/figures

\definecolor{f1blue}{RGB}{76, 114, 176}
\definecolor{aucgreen}{RGB}{85, 168, 104}
\definecolor{aporange}{RGB}{221, 132, 82}

\pgfplotsset{compat=1.18}
\usetikzlibrary{arrows.meta, calc, positioning, fit}

\copyrightyear{2026}
\acmYear{2026}
\setcopyright{cc}
\setcctype{by}

\acmConference[CIKM '26]
{Proceedings of the 35th ACM International Conference on Information and Knowledge Management}
{November 07--11, 2026}
{Rome, Italy}

\acmBooktitle{Proceedings of the 35th ACM International Conference on Information and Knowledge Management (CIKM '26), November 07--11, 2026, Rome, Italy}

\acmDOI{10.1145/3799682.3840971}
\acmISBN{979-8-4007-2539-5/2026/11}

\title{CSC: Calibrated Simplicity for Conflict-Aware Social Bot Detection in the LLM Era}

\author{Yipeng Qian}
\authornote{Yipeng Qian and Pengjie Zhao contributed equally to this work.}
\affiliation{%
  \institution{Minzu University of China}
  \city{Beijing}
  \country{China}
}
\email{24160092@muc.edu.cn }

\author{Pengjie Zhao}
\authornotemark[1]
\affiliation{%
  \institution{Minzu University of China}
  \city{Beijing}
  \country{China}
}
\email{24160162@muc.edu.cn}

\author{Chaoxi Niu}
\authornote{Corresponding author.}
\affiliation{%
  \institution{ City University of Macau}
  \city{Macau}
  \country{China}
}
\email{niuchaoxi@gmail.com}

\renewcommand{\shortauthors}{Qian et al.}



\begin{document}

\begin{abstract}
Social bot detection is essential for protecting online platforms from misinformation amplification, coordinated manipulation, and distorted public discourse. However, large language models have made social bots much harder to detect from text alone because semantic camouflage is now cheap, fluent, and scalable. The resulting challenge is modality conflict: an account may look human-like in semantics while remaining suspicious in graph structure, profile attributes, or cross-modal consistency. Recent graph-based detectors tackle this limitation by adding graph-side complexity, such as sparse prototype selection, adaptive gating, or architecture-specific control logic, yet our experiments suggest that complexity alone is not the most reliable way to resolve such conflict.

We therefore propose \textsc{CSC}, a calibrated-simplicity framework for conflict-aware LLM-era social bot detection. The framework combines three design choices: a simplified prototype-guided graph expert that retains useful structural biases while removing unstable graph-side heuristics, calibrated simplex-constrained fusion that aligns heterogeneous confidence spaces before late fusion, and a lightweight inconsistency expert that models cross-modal disagreement. Experiments on TwiBot-22, TwiBot-20, and MGStBot-large show that \textsc{CSC} improves calibrated operating-point decision quality while remaining competitive across external benchmarks. Further analyses show that calibration improves confidence reliability, the inconsistency expert mainly provides localized corrections in high-conflict or near-threshold regions, and simplified graph-side control yields a better stability-cost trade-off. A targeted semantic-camouflage stress test further shows that replacing selected bot text with matched human text sharply degrades the standalone text expert while leaving graph and fused evidence stable on a balanced challenge set. Overall, these results suggest that, when semantic evidence is easy to camouflage, robust bot detection benefits from preserving reliable structural signals and calibrating conflicting modalities before fusion, rather than continually increasing graph-side complexity.
\end{abstract}
% Across TwiBot-22, TwiBot-20 under the MGTAB-style protocol, and MGStBot-large, \textsc{CSC} achieves 0.5956 F1 on TwiBot-22, 0.8828 F1 on TwiBot-20, and 0.8532 mean F1 on MGStBot-large under a unified three-seed evaluation. Taken together with graph ablations, conflict-region analysis, and efficiency measurements, these results support a focused claim: 

\maketitle

\section{Introduction}
Social bot detection aims to identify automated or semi-automated accounts on social media platforms. 
It is an important task because social bots can amplify misinformation, manipulate online discussions, distort public opinion, and undermine the trustworthiness of social platforms~\citep{ferrara2016rise,subrahmanian2016darpa,bessi2016social,shao2018spread,vosoughi2018spread}.
Over the past decade, social bot detection has developed into an arms race between automated accounts and the systems designed to detect them~\citep{cresci2017paradigm,varol2017online,yang2020scalable}.
Earlier detectors relied on profile statistics, temporal patterns, and textual cues~\citep{kantepe2017preprocessing,varol2017online,cresci2017paradigm}, while recent systems increasingly incorporate pretrained language encoders and graph neural networks to capture semantic and relational evidence~\citep{liu2019roberta,feng2021botrgcn,feng2022rgt,feng2022twibot22}. 
The rapid development of large language models (LLMs) further escalates this arms race. 
Modern bots can now generate fluent profiles, coherent posts, and plausible conversational behavior at low cost and large scale, making purely semantic cues less reliable than before~\citep{feng2024what,qiao2025botsim,wang2026tracebot}.

\begin{figure}[t]
\centering
\includegraphics[width=\columnwidth]{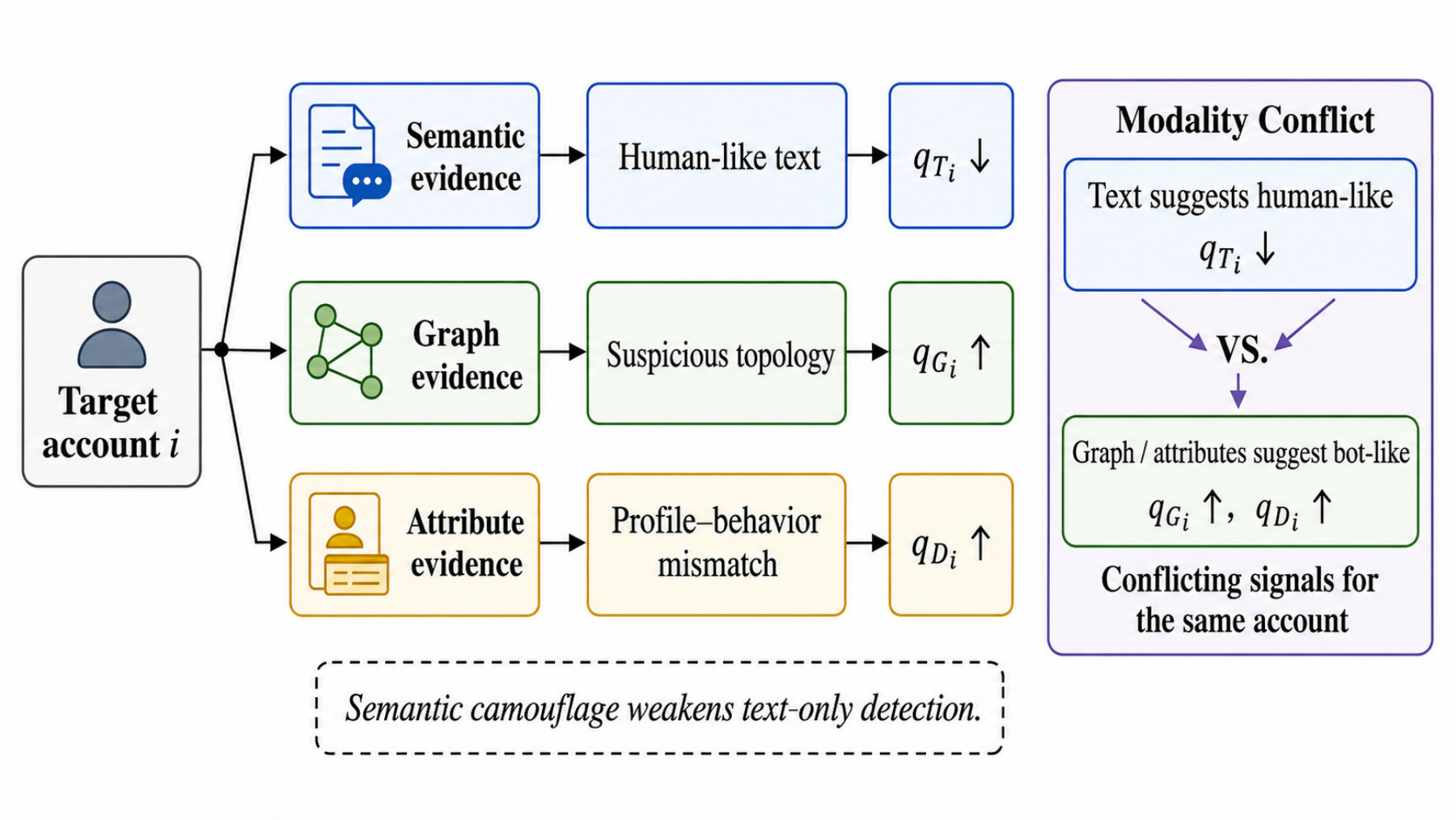}
\caption{Modality conflict in the LLM era. 
A target account may receive different signals from different evidence sources: human-like language leads the text-side expert to assign a low bot score $q_i^T$, while suspicious topology and profile--behavior mismatch lead the graph- and attribute-side experts to assign high bot scores $q_i^G$ and~$q_i^D$. 
Here $q_i^T$, $q_i^G$, and $q_i^D$ denote the text-, graph-, and attribute-side expert scores for account $i$, respectively. 
This conflict motivates confidence calibration and inconsistency modeling.}
\Description{A conceptual diagram showing a target account with conflicting evidence: text appears human-like, while graph topology and profile-behavior consistency appear suspicious.}
\label{fig:modality-conflict}
\end{figure}

However, this shift creates a modality-conflict problem as shown in Fig.~\ref{fig:modality-conflict}. 
An account may appear human-like in language while still exposing suspicious patterns in graph structure, profile attributes, or cross-modal consistency. 
Textual abnormality, once a useful signal for bot detection, can now be deliberately weakened by semantic camouflage. 
At the same time, non-textual evidence such as unusual interaction topology, suspicious coordination, and profile-behavior inconsistency remains informative. 
The central challenge is therefore no longer only how to extract stronger single-modal features, but how to make reliable decisions when heterogeneous evidence disagrees.

Existing approaches only partially address this challenge. 
Text-centered detectors remain vulnerable when bots imitate human language, while graph-based detectors provide a complementary view by modeling relational and heterogeneous structure~\citep{feng2021botrgcn,feng2022rgt,hu2020hgt,lv2021simplehgn}. 
However, increasing graph-side complexity through additional prototype selection, adaptive gating, or architecture-specific control logic does not directly resolve modality conflict. 
These mechanisms may increase model expressiveness, but they also introduce more design choices and make the source of the final improvement less clear. Such modules may improve representation capacity, but they also make it harder to determine whether gains come from better graph reasoning or from extra tuning choices.
Naive multimodal fusion has a similar limitation, i.e., directly concatenating features or averaging branch scores assumes that different experts produce comparable confidence values, which is rarely guaranteed in practice~\citep{atrey2010multimodal,ngiam2011multimodal,baltrusaitis2019multimodal,guo2017calibration}.

Motivated by this observation, we propose \textsc{CSC}, a conflict-aware social bot detection framework built around calibrated simplicity. 
Instead of continually adding graph-side control logic, \textsc{CSC} preserves structural biases that consistently help, aligns heterogeneous confidence scores before fusion, and explicitly models disagreement among modalities. Specifically, CSC contains three components: a simplified prototype-guided graph expert for structural evidence, calibrated simplex fusion for validation-only confidence alignment, and a lightweight inconsistency expert for disagreement-aware correction.

% \ncx{add a paragraph to briefly present the details of each module and the motivation.}
The first component, the simplified prototype-guided graph expert, is motivated by the observation that graph structure remains informative when textual evidence is semantically camouflaged. 
It keeps dual-view encoding, dense multi-prototype discrimination, and fixed-strength residual correction, while avoiding additional graph-side controls that make the model harder to interpret. 
The second component, calibrated simplex fusion, addresses the confidence mismatch among heterogeneous experts by mapping their outputs into a comparable probability space before fusion. 
The third component, the inconsistency expert, uses cross-branch disagreement features to capture cases where text, graph, and attribute evidence point in different directions. 
Together, these modules allow the detector to preserve useful structural evidence, align heterogeneous confidence scores, and use cross-modal disagreement as an auxiliary signal for localized correction.
Rather than arguing that graph simplification is universally preferable or that generic ensembling is sufficient, we view LLM-era social bot detection as conflict-aware evidence integration. 
\textsc{CSC} therefore keeps reliable structural biases, calibrates heterogeneous confidence scores, and uses cross-modal disagreement as a lightweight correction signal.

Our contributions are summarized as follows:
\begin{itemize}    
    \item We formulate LLM-era social bot detection as a modality-conflict problem, where fluent language can hide automated behavior while graph structure, profile attributes, or cross-modal consistency still expose suspicious patterns.

    \item We propose \textsc{CSC}, a conflict-aware detection framework that combines a simplified prototype-guided graph expert, validation-only calibrated simplex fusion, and a lightweight inconsistency expert for disagreement-aware correction.

    \item We conduct a protocol-conscious empirical study across TwiBot-22, TwiBot-20, and MGStBot-large, showing that calibrated evidence integration improves operating-point decision quality under modality conflict while remaining competitive across external benchmarks.
\end{itemize}

\section{Related Work}

\subsection{Social Bot Detection}

Social bot detection has moved from feature-based pipelines to
learning-based systems that combine semantic, behavioral, and
relational evidence. Early methods used profile statistics, temporal
patterns, and hand-crafted behavioral descriptors to separate
automated accounts from human users~\cite{cresci2017paradigm,kantepe2017preprocessing,sayyadiharikandeh2020detection,varol2017online,yang2020scalable}. 
Later work introduced stronger tabular models and pretrained language
encoders for dense profile and text representations~\cite{ke2017lightgbm,liu2019roberta}. 
Large-scale benchmarks such as TwiBot-20, TwiBot-22, and MGTAB
further showed that realistic detection cannot be reduced to text or
profile classification, since social context and graph structure often
provide crucial evidence for coordinated behavior~\cite{feng2021twibot20,feng2022twibot22,shi2023mgtab}. 
LLMs make this setting more challenging. Feng et al.~\cite{feng2024what}
show that LLMs can help social bot detection, but can also be used to
manipulate account content and weaken existing detectors. Recent
LLM-powered bot simulation and LLM-driven bot datasets also suggest
that future detectors need to account for stronger semantic camouflage
and cross-platform distribution shifts~\cite{qiao2025botsim,wang2026tracebot}. 
Our paper focuses on one consequence of this shift: once text becomes
easier to imitate, robust detection depends on how a model handles
conflicts between semantic, structural, and attribute evidence.

Recently, graph neural networks have been widely used for
topology-aware social bot detection. BotRGCN~\cite{feng2021botrgcn}
models Twitter interactions with relational graph convolution, and
RGT~\cite{feng2022rgt} further considers relation heterogeneity and
influence heterogeneity through relational graph transformers. More
recent work explores community-aware and heterophily-aware graph
reasoning, such as BotCF~\cite{liu2025botcf} and
BotSCL~\cite{wu2023botscl}. General graph architectures, including
GraphSAGE~\cite{hamilton2017graphsage}, GCN~\cite{kipf2017gcn},
GAT~\cite{velickovic2018gat}, HGT~\cite{hu2020hgt}, and
Simple-HGN~\cite{lv2021simplehgn}, are also commonly used as
baselines for measuring the value of task-specific graph designs.
BotCF models multimodal and community interactions implicitly at the representation level. 
In contrast, \textsc{CSC} focuses on decision-level confidence alignment by explicitly calibrating heterogeneous expert outputs before constrained fusion; the two mechanisms are therefore complementary rather than mutually exclusive.
These studies establish graph structure as an important signal for
social bot detection, especially when textual evidence becomes less
reliable. However, most graph-enhanced detectors focus on improving
graph-side representation capacity through relation modeling,
heterogeneity modeling, community features, or contrastive objectives.
Such designs strengthen structural reasoning, but they do not directly
address how graph, text, profile, and structured evidence should be
calibrated and fused when their predictions conflict. Our work is
closest to graph-enhanced bot detection, but shifts the emphasis from
adding graph mechanisms to retaining reliable structural biases and
integrating heterogeneous evidence in a calibrated, conflict-aware
manner.

\subsection{Heterogeneous Fusion and Calibration}

Practical bot detection systems often combine text, profile attributes,
behavior statistics, and graph structure. The fusion step, however,
is often treated as a simple concatenation or score-averaging problem
\cite{atrey2010multimodal,baltrusaitis2019multimodal,ngiam2011multimodal}.
This can be fragile because different branches may produce scores on
different confidence scales \cite{guo2017calibration,niculescu2005predicting,platt1999probabilistic}.
For example, a semantic expert can become overconfident in fluent-generated
text, while a graph expert may provide weaker but more reliable evidence
for suspicious relational patterns.

Calibration and residual modeling are useful tools for this setting.
Probability calibration has long been studied for mapping raw classifier
outputs to more reliable confidence estimates
\cite{platt1999probabilistic,lin2007platt,niculescu2005predicting,guo2017calibration,zadrozny2001calibrated}.
In attributed-network anomaly detection, reconstruction and residual
signals have been used to capture cases that are not fully explained by
smooth representations alone \cite{ding2019dominant,fan2020anomalydae}.
In social bot detection, recent work also suggests that robustness depends
on modeling nontrivial neighborhood structure rather than only strengthening
semantic encoders \cite{wu2023botscl}.

Our work builds on these insights but focuses on a different problem:
how to make a final decision when calibrated experts disagree. Rather than
treating fusion as generic ensembling, \textsc{CSC} first aligns
heterogeneous confidence scores, then combines them under a validation-only
simplex constraint, and finally adds a lightweight inconsistency expert to
capture cross-modal disagreement. This allows the model to use disagreement
as conflict evidence instead of simply averaging it away.

\section{Problem Formulation}

We formulate social bot detection as binary node classification on a heterogeneous social graph. 
Let $G=(V, E, R)$ denote a directed heterogeneous graph, where $V$ is the set of accounts, $E$ is the set of edges, and $R$ denotes relation types such as following and follower links. 
Each account $i \in V$ is associated with heterogeneous attributes:
\[
x_i = \{x_i^{\text{text}}, x_i^{\text{profile}}, x_i^{\text{struct}}\},
\]
where $x_i^{\text{text}}$ denotes language-based content such as posts and profile descriptions, $x_i^{\text{profile}}$ denotes non-textual account metadata such as account age, verification status, follower-related counts, and other profile statistics, and $x_i^{\text{struct}}$ denotes structured behavioral or interaction features. 
This separation allows semantic evidence, account-level metadata, and behavioral evidence to be modeled as distinct information sources.

Given labels $y_i \in \{0,1\}$ for a subset of accounts, where $1$ denotes a bot and $0$ denotes a human account, the goal is to learn a scoring function
\[
f_\theta(i;G,X) \in [0,1],
\]
where $X=\{x_i\}_{i\in V}$ denotes the account features and a larger score indicates a higher probability that account $i$ is automated.

The setting considered in this paper is characterized by \emph{modality conflict}. 
In LLM-era social bot detection, textual evidence may become less reliable because automated accounts can generate fluent and human-like content. 
At the same time, graph topology, profile metadata, and structured behavioral signals may still expose suspicious patterns. 
As a result, different modalities may support different predictions for the same account: text may suggest a human-like account, while graph, profile, or structured-feature evidence suggests bot-like behavior.

This conflict changes the design objective of a detector. 
A robust model should not only learn strong modality-specific representations, but also decide how to combine heterogeneous evidence when the corresponding experts disagree. 
This motivates \textsc{CSC}, which preserves useful structural evidence, calibrates heterogeneous confidence scores, and explicitly models disagreement before making the final prediction.
% \section{Method}

% In this section, we present \textsc{CSC}, a staged multi-expert framework for LLM-era social bot detection.
% The central idea is to treat cross-modal disagreement as an explicit source of evidence rather than as noise to be averaged out.
% As shown in Figure~\ref{fig:overview}, \textsc{CSC} consists of three key components: 
% a simplified prototype-guided graph expert, a confidence-aligned multi-expert calibration module, and a lightweight inconsistency expert.
% The final prediction is obtained through simplex-constrained fusion over calibrated expert outputs.
\section{Method}

In this section, we present \textsc{CSC}, a staged multi-expert framework for LLM-era social bot detection. 
The central idea is to calibrate heterogeneous expert scores before fusion and use cross-modal disagreement as an auxiliary cue when the experts provide conflicting evidence. 
As shown in Figure~\ref{fig:framework}, \textsc{CSC} first obtains modality-specific expert scores from graph, text, profile, and structured evidence, then calibrates these scores and augments them with a lightweight inconsistency expert before simplex-constrained fusion. 
The final prediction is therefore made from calibrated and conflict-aware evidence rather than from raw branch outputs alone.
% \zpj{concept?:The final prediction is made after raw expert scores are calibrated, combined under simplex constraints, and supplemented with disagreement features.}

\subsection{Overview}

The framework separates the final decision into three roles: structural evidence modeling, confidence alignment, and disagreement-aware correction.
Given an account $i$, \textsc{CSC} first obtains predictions from a simplified graph expert and a small set of non-graph experts derived from text, profile, and structured features. 
Let
\[
\mathcal{M}=\{G,P,T,D\}
\]
denote the base expert set, where $G$ is the graph expert and $P$, $T$, and $D$ denote the profile, text, and structured-feature experts, respectively. 
Each expert outputs a raw score $p_i^{(m)}$, which is later calibrated on the validation split to produce a comparable confidence score $q_i^{(m)}$. 
The calibrated scores are combined through simplex-constrained fusion. 
In addition, a lightweight inconsistency expert $C$ is built from cross-branch disagreement features and added to the final fusion stage.

Compared with heavier prototype-guided variants, \textsc{CSC} makes two deliberate simplifications in the graph branch. First, dense prototype aggregation replaces sparse prototype selection, keeping the ranking signal smooth during optimization. Second, a fixed residual coefficient replaces sample-wise gating, avoiding another unstable decision layer. These choices reduce graph-side control logic while preserving useful structural biases, making the graph expert easier to calibrate and integrate with non-graph experts.

\begin{figure*}[t]
\centering
\includegraphics[width=0.94\textwidth]{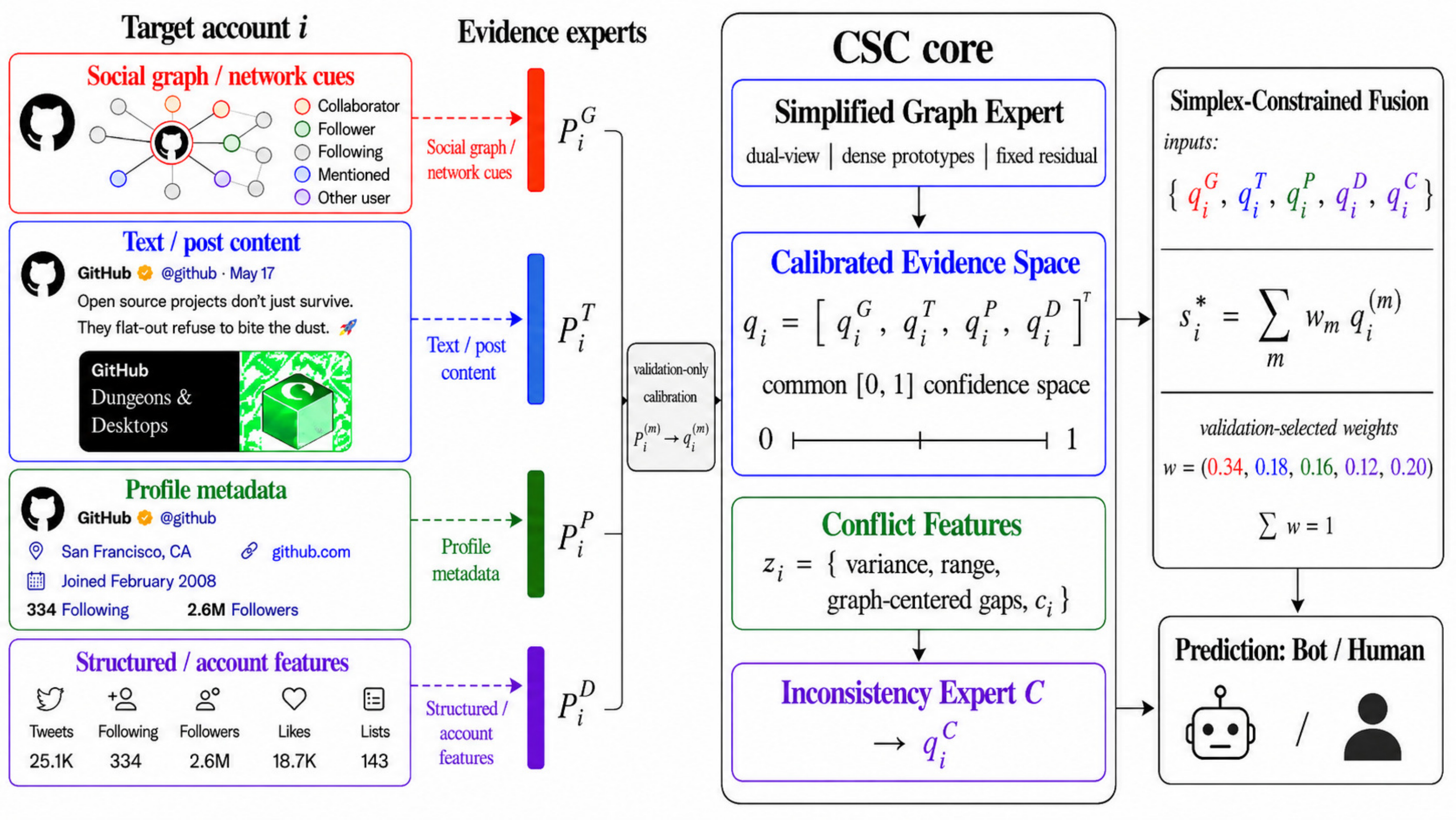}
\caption{Overview of \textsc{CSC}. The framework extracts graph, text, profile, and structured evidence from a target account and obtains modality-specific expert scores. These scores are independently calibrated using validation-only Platt scaling. A lightweight inconsistency expert captures cross-branch disagreement, and all calibrated outputs are combined through simplex-constrained fusion to produce the final bot-detection decision.}
\Description{A publication-quality schematic of CSC showing a target account feeding multiple evidence experts, a validation-only calibration stage, a lightweight inconsistency expert, and a final simplex-constrained fusion layer that outputs the bot-detection prediction.}
\label{fig:framework}
\end{figure*}

Figure~\ref{fig:framework} summarizes how the graph expert, calibration module, inconsistency expert, and simplex-constrained fusion layer are connected in the full pipeline.

\subsection{Simplified Prototype-Guided Graph Expert}

We first describe the graph expert, which provides the structural evidence used by the subsequent calibrated fusion stage. 
Under modality conflict, semantic signals may become unreliable due to fluent text generation, while relational patterns and local topology can still expose suspicious behavior. 
The goal of this branch is therefore not to maximize graph-side complexity, but to retain structural biases that are consistently useful in this setting. 
To this end, we adopt a simplified prototype-guided graph expert with three components: dual-view encoding, dense multi-prototype discrimination, and fixed-strength residual correction.

\paragraph{Dual-view encoding.}
For each node $i$, we construct a graph representation by combining neighborhood aggregation with a self-feature pathway:
\begin{equation}
h_i =
\operatorname{LN}\!\left(
W_f
\left[
\operatorname{SAGE}(x_i,\mathcal{N}_i)
\,\Vert\,
\operatorname{MLP}(x_i)
\right]
+b_f
\right),
\end{equation}
where $\mathcal{N}_i$ denotes the neighborhood of node $i$, $\Vert$ denotes concatenation, and $\operatorname{LN}$ denotes layer normalization. 
The neighborhood branch captures relational context, while the self-feature branch preserves account-specific information that may otherwise be diluted by aggregation.

\paragraph{Dense multi-prototype discrimination.}
Instead of representing each class with a single centroid, we introduce learnable positive and negative prototype sets $P^+, P^- \in \mathbb{R}^{K\times d}$. 
For each node representation $h_i$, we compute temperature-scaled cosine similarities to all prototypes:
\begin{equation}
s_{ik}^{+}
=
\frac{\langle \hat{h}_i,\hat{p}_k^{+}\rangle}{\tau_p},
\qquad
s_{ik}^{-}
=
\frac{\langle \hat{h}_i,\hat{p}_k^{-}\rangle}{\tau_p},
\end{equation}
where $\hat{h}_i=h_i/\|h_i\|_2$, $\hat{p}_k^{\pm}=p_k^{\pm}/\|p_k^{\pm}\|_2$, and $\tau_p>0$ is a temperature parameter. 
We aggregate all similarities using a dense log-sum-exp rule:
\begin{equation}
\ell_i^{\mathrm{proto}}
=
\gamma_p
\left[
\log\sum_{k=1}^{K}\exp(s_{ik}^{+})
-
\log\sum_{k=1}^{K}\exp(s_{ik}^{-})
\right],
\end{equation}
where $\gamma_p$ controls the scale of prototype evidence. 
This formulation preserves global class-level ranking information while avoiding the hard selection behavior introduced by sparse top-$r$ prototype truncation.

\paragraph{Fixed-strength residual correction.}
Prototype evidence captures global class-level structure but may be insufficient near decision boundaries. 
We therefore construct a compact prototype-geometry summary:
\begin{equation}
d_i =
\left[
\max_k s_{ik}^{+},
\max_k s_{ik}^{-},
\mu_i^{+},
\mu_i^{-},
\ell_i^{\mathrm{proto}}
\right],
\end{equation}
where
\begin{equation}
\mu_i^{+}=\frac{1}{K}\sum_{k=1}^{K}s_{ik}^{+},
\qquad
\mu_i^{-}=\frac{1}{K}\sum_{k=1}^{K}s_{ik}^{-}.
\end{equation}
A lightweight residual scorer produces a correction term:
\begin{equation}
\Delta_i=g_{\phi}(d_i).
\end{equation}
The final graph logit is
\begin{equation}
\ell_i^{G}
=
\ell_i^{\mathrm{proto}}
+
\eta_0\Delta_i,
\end{equation}
where $\eta_0$ is a fixed coefficient. 
The graph probability is then
\begin{equation}
p_i^{G}=\sigma(\ell_i^{G}).
\end{equation}
Using a fixed correction strength avoids introducing another sample-wise gating mechanism, keeping the graph branch compact and stable on imbalanced benchmarks.

\subsection{Calibrated Simplex Fusion}

The graph expert provides structural evidence, while the non-graph
experts capture complementary semantic, profile, and structured
signals. 
For the non-graph modalities, we use separate predictors for profile metadata, textual content, and structured behavioral features. Their outputs are
\[
p_i^P=\sigma\!\left(f_P(x_i^{\mathrm{profile}})\right),\qquad
p_i^T=\sigma\!\left(f_T(x_i^{\mathrm{text}})\right),\qquad
p_i^D=\sigma\!\left(f_D(x_i^{\mathrm{struct}})\right).
\]
The profile expert captures account-level metadata, the text expert models semantic evidence that may be vulnerable to LLM-based camouflage, and the structured-feature expert captures non-text behavioral and interaction statistics. Together with the graph score $p_i^G$, these form the raw expert outputs $\{p_i^{(m)}\}_{m\in\mathcal{M}}$, where $\mathcal{M}=\{G,P,T,D\}$.

However, raw expert outputs are often not directly comparable because different experts operate in different confidence spaces. 
For example, a semantic expert may become overconfident on fluent generated content, while a graph expert may produce less confident but more reliable scores for suspicious relational patterns. 
Directly combining such raw outputs can therefore lead to unstable or misleading decisions.

To address this issue, we calibrate each expert independently using validation-only Platt scaling~\citep{platt1999probabilistic,lin2007platt,guo2017calibration}. 
For each expert $m\in\mathcal{M}$, the raw score $p_i^{(m)}$ is transformed into a calibrated confidence score:
\begin{equation}
q_i^{(m)}
=
\sigma\left(
a_m \operatorname{logit}(p_i^{(m)}) + b_m
\right),
\end{equation}
where $(a_m,b_m)$ are fitted on validation predictions by minimizing binary negative log-likelihood. 
This step maps heterogeneous expert outputs into a comparable confidence space without modifying the internal structure of each expert.

After calibration, we combine the experts using simplex-constrained fusion:
\begin{equation}
s_i
=
\sum_{m\in\mathcal{M}} w_m q_i^{(m)},
\qquad
\sum_{m\in\mathcal{M}} w_m=1,\quad
w_m\ge 0.
\end{equation}
The fusion weights and the operating threshold are selected jointly on the validation split under the simplex constraint. 
Specifically, when calibrated base fusion is used, we choose the weights and threshold by maximizing validation F1:
\[
(w^\star,\tau^\star)
=
\arg\max_{w\in\Delta^{|\mathcal{M}|},\,\tau\in[0,1]}
\mathrm{F1}_{\mathrm{val}}
\left(
\mathbf{1}\left[
\sum_{m\in\mathcal{M}} w_m q_i^{(m)} \ge \tau
\right],
y_i
\right),
\]
where $\Delta^{|\mathcal{M}|}=\{w:\sum_{m\in\mathcal{M}}w_m=1,\;w_m\ge0\}$ denotes the probability simplex. 
In implementation, candidate weights are searched under the simplex constraint, and the decision threshold is scanned on the validation split. 
The same validation-only search rule is used for all fusion variants, and no test labels are used for calibration, weight selection, or threshold tuning. 
The resulting base-fusion score is
\[
s_i^\star=\sum_{m\in\mathcal{M}} w_m^\star q_i^{(m)}.
\]

After the inconsistency expert is introduced in Section~\ref{sec:inconsistency}, the same validation-only simplex rule is applied to the augmented expert set.
% \ncx{how is $w_m$ determined?}

\subsection{Inconsistency Expert}
\label{sec:inconsistency}

Calibrated simplex fusion aligns expert confidence scores and provides an interpretable decision rule. 
However, a convex combination alone can obscure different forms of expert disagreement. 
For example, one account may receive moderately high bot scores from all experts, while another may receive a low text-side score but high graph- and attribute-side scores. 
Although their fused scores can be similar, the latter case reflects a stronger cross-modal conflict. 
This distinction is important in LLM-era bot detection, where disagreement between human-like semantics and suspicious non-textual evidence can itself be informative.

To capture such signals, we introduce a lightweight inconsistency expert. 
It uses score-level disagreement statistics together with compact profile--behavior consistency features. 
For each account $i$, let
$\mathbf{c}_i =
\phi_c\!\left(x_i^{\mathrm{profile}},x_i^{\mathrm{struct}}\right)$
denote a low-dimensional consistency vector computed from profile metadata and structured behavioral features. 
The function $\phi_c(\cdot)$ extracts normalized indicators of profile--behavior consistency, such as profile completeness, activity-density ratios, follower--following imbalance, and profile--behavior mismatch patterns.

We then construct a meta-feature vector from calibrated expert scores, cross-branch disagreement statistics, and consistency features:
\begin{equation}
\begin{aligned}
z_i=\operatorname{concat}\Big(
& q_i^G, q_i^P, q_i^T, q_i^D,
\operatorname{Var}(q_i^{(\cdot)}),
\operatorname{Range}(q_i^{(\cdot)}), \\
& |q_i^G-q_i^T|,
|q_i^G-q_i^P|,
|q_i^G-q_i^D|,
\mathbf{c}_i
\Big),
\end{aligned}
\end{equation}
where $q_i^G$ is the calibrated graph score, and $q_i^P$, $q_i^T$, and $q_i^D$ are the calibrated profile, text, and structured-feature expert scores, respectively. 
The variance, range, and graph-centered gaps summarize score-level disagreement, while $\mathbf{c}_i$ contributes profile--behavior consistency evidence that is not fully represented by expert probabilities alone.
Importantly, \textsc{CSC} does not assume that larger cross-modal disagreement necessarily indicates bot behavior. 
Instead, disagreement is interpreted jointly with calibrated expert scores and profile--behavior consistency features, serving as an auxiliary diagnostic cue rather than sufficient evidence for a bot prediction.
The inconsistency expert maps this meta-feature vector to a raw conflict-aware score:
\begin{equation}
r_i = h_{\psi}(z_i),
\end{equation}
where $h_{\psi}$ is a lightweight tabular classifier. 
We calibrate this score in the same validation-only manner as the base experts:
\begin{equation}
q_i^C
=
\sigma\left(
a_C\operatorname{logit}(r_i)+b_C
\right).
\end{equation}
We denote the calibrated inconsistency expert by $C$, whose output for account $i$ is $q_i^C$. 
The inconsistency expert is then added to the base expert set:
\begin{equation}
\mathcal{M}^{+}
=
\mathcal{M}\cup\{C\}.
\end{equation}

The final fused score becomes
\begin{equation}
s_i^\star
=
\sum_{m\in\mathcal{M}^{+}} w_m q_i^{(m)},
\qquad
\sum_{m\in\mathcal{M}^{+}} w_m=1,\quad
w_m\ge 0,
\end{equation}
and the prediction is
\begin{equation}
\hat{y}_i=\mathbf{1}(s_i^\star\ge \tau^\star).
\end{equation}

The inconsistency expert is not designed as a standalone detector or a replacement for calibrated fusion. 
Rather, it provides a lightweight correction signal for cases where disagreement patterns are compressed by a single convex fusion score. 
In this way, \textsc{CSC} treats cross-modal disagreement as localized evidence for decision correction instead of averaging it away.
\subsection{Training and Inference}

The graph expert is trained with supervised binary classification and prototype-level supervision. 
Let $p_i^G=\sigma(\ell_i^G)$ denote the graph-branch probability for node $i$, and let $y_i\in\{0,1\}$ denote its label. 
The main graph classification loss is a class-weighted binary cross-entropy:
\begin{equation}
\mathcal{L}^{G}_{\mathrm{cls}}
=
-\frac{1}{|\mathcal{V}_{\mathrm{train}}|}
\sum_{i\in\mathcal{V}_{\mathrm{train}}}
\left[
\omega_1 y_i \log p_i^G
+
\omega_0(1-y_i)\log(1-p_i^G)
\right],
\end{equation}
where $\omega_1$ and $\omega_0$ are class weights for bot and human accounts, respectively.

To keep the prototype branch itself discriminative, we also apply supervision to the prototype-only probability $p_i^{\mathrm{proto}}=\sigma(\ell_i^{\mathrm{proto}})$:
\begin{align}
\mathcal{L}_{\mathrm{proto}}
= & \\
 -\frac{1}{|\mathcal{V}_{\mathrm{train}}|}&
\sum_{i\in\mathcal{V}_{\mathrm{train}}}
\left[
\omega_1 y_i \log p_i^{\mathrm{proto}}
+
\omega_0(1-y_i)\log(1-p_i^{\mathrm{proto}})
\right].
\end{align}
The graph objective is
\begin{equation}
\mathcal{L}_{G}
=
\mathcal{L}^{G}_{\mathrm{cls}}
+
\lambda_{\mathrm{proto}}\mathcal{L}_{\mathrm{proto}},
\end{equation}
where $\lambda_{\mathrm{proto}}$ controls the strength of prototype supervision.

The non-graph experts are trained independently on the same training split with their own supervised objectives. 
After the base experts are trained, their calibration parameters are fitted using validation predictions only. 
The inconsistency expert is then trained from frozen, calibrated expert outputs and attribute-consistency features. 
Finally, the simplex weights and the decision threshold $\tau^\star$ are selected on the validation split.

This staged protocol is important for preventing information leakage. 
No test labels are used for calibration, simplex-weight selection, or threshold tuning. 
At inference time, \textsc{CSC} computes the raw expert scores, calibrates them with validation-fitted calibration parameters, constructs the inconsistency score, applies the validation-selected simplex weights, and predicts the label using the validation-selected threshold.

% The full training procedure is staged as follows. 
% \ncx{using algorithm style instead.}
Algorithm~\ref{alg:s3fusion} summarizes the staged training and inference protocol.

\begin{algorithm}[!htbp]
\caption{Training and Inference of \textsc{CSC}}
\label{alg:s3fusion}
\small
\begin{algorithmic}[1]
\Require Social graph $G$, account features $X$, training split $\mathcal{V}_{\mathrm{train}}$, validation split $\mathcal{V}_{\mathrm{val}}$, test account $i$, base expert set $\mathcal{M}=\{G,P,T,D\}$
\Ensure Predicted label $\hat{y}_i$

\State Train graph expert $G$ on $\mathcal{V}_{\mathrm{train}}$ with $\mathcal{L}_{G}$.
\State Train non-graph experts $P$, $T$, and $D$ independently on $\mathcal{V}_{\mathrm{train}}$.
\State Freeze all base experts and obtain validation predictions.
\For{each base expert $m\in\mathcal{M}$}
    \State Fit calibration parameters $(a_m,b_m)$ on $\mathcal{V}_{\mathrm{val}}$.
    \State Compute calibrated scores $q^{(m)}$.
\EndFor
\State Construct inconsistency features $z$ from calibrated scores and conflict statistics.
\State Train and calibrate the inconsistency expert $C$ using frozen base-expert outputs.
\State Form the augmented expert set $\mathcal{M}^{+}=\mathcal{M}\cup\{C\}$.
\State Select simplex weights $w^\star$ and threshold $\tau^\star$ on $\mathcal{V}_{\mathrm{val}}$ by maximizing validation F1.

\Statex
\State \textbf{Inference:}
\State Compute calibrated scores $\{q_i^{(m)}\}_{m\in\mathcal{M}^{+}}$ for test account $i$.
\State Compute final score $s_i^\star=\sum_{m\in\mathcal{M}^{+}}w_m^\star q_i^{(m)}$.
\State \Return $\hat{y}_i=\mathbf{1}(s_i^\star\ge\tau^\star)$.
\end{algorithmic}
\end{algorithm}

\section{Experiments}
\subsection{Datasets and Evaluation Metrics}

We evaluate \textsc{CSC} on three main bot-detection benchmarks. 
TwiBot-22 is the primary benchmark for severe modality conflict~\cite{feng2022twibot22}. 
TwiBot-20 is reported under the standardized comparison protocol summarized by MGTAB~\cite{feng2021twibot20,shi2023mgtab}. 
MGStBot-large is used as an external benchmark.

For binary bot detection, we report accuracy, precision, recall, F1, and AUC when available. 
F1 and AUC serve complementary roles: F1 reflects the operating-point quality of the final detector, while AUC measures threshold-free ranking performance. 
For calibration diagnostics, we additionally report expected calibration error (ECE), Brier score, and negative log-likelihood (NLL).

We additionally include a reduced BotSim-24 sanity check~\cite{qiao2025botsim} to examine whether the LLM-era motivation is consistent with public LLM-driven bot data. 
Because BotSim-24 is Reddit-based and not schema-compatible with the TwiBot-style pipeline, we use it only as an auxiliary external validation rather than as a main final-model comparison.
\subsection{Baselines and Experimental Protocol}

The comparison pool covers graph, semantic, and tabular baselines. 
It includes task-specific graph detectors such as BotRGCN~\cite{feng2021botrgcn}, RGT~\cite{feng2022rgt}, BotSCL~\cite{wu2023botscl}, and a BotCF-style reproduction~\cite{liu2025botcf}; generic graph backbones such as GraphSAGE~\cite{hamilton2017graphsage}, GCN~\cite{kipf2017gcn}, GAT~\cite{velickovic2018gat}, HGT~\cite{hu2020hgt}, and Simple-HGN~\cite{lv2021simplehgn}; and non-graph baselines such as Kantepe-LR~\cite{kantepe2017preprocessing}, LightGBM~\cite{ke2017lightgbm}, and RoBERTa-direct~\cite{liu2019roberta}. 
The main comparison table reports final-model performance, while later diagnostic tables separate graph-branch choices, calibration effects, fusion choices, threshold effects, and conflict-region behavior.

Because the baselines are drawn from different released implementations and reporting protocols, we organize the evaluation into three comparison scopes. 
First, the TwiBot-22 local comparison evaluates representative graph, semantic, and non-graph baselines under a shared local pipeline. 
Second, the TwiBot-20 comparison follows the MGTAB-style protocol to support comparison with established benchmark reports. 
Third, MGStBot-large evaluates external transfer, with official large-graph baselines reproduced under the local benchmark setting. 
Table~\ref{tab:main-results} reports these scopes together for compactness; row grouping, caption notes, and ``--'' entries indicate which methods were not evaluated under a given benchmark.
Accordingly, unavailable entries are not used for direct pairwise superiority claims, and our mechanism-level conclusions are drawn from protocol-controlled local comparisons and diagnostics.
For \textsc{CSC}, calibration functions, simplex weights, and decision thresholds are fitted or selected on the validation split only. 
This validation-only rule is part of the evaluation protocol because hard benchmarks such as TwiBot-22 are sensitive to the operating point of the final detector. 
Unless otherwise stated, the graph branch, non-graph branches, and fusion stage use the same train/validation/test split within each benchmark configuration. 
The final fused predictions on TwiBot-20 are summarized with bootstrap estimation over 3000 resamples to obtain the reported uncertainty ranges.

We include a protocol-matched BotCF-style reproduction as a recent community-aware baseline~\cite{liu2025botcf}. 
Because the official implementation was not available for this evaluation and the original DANMF community extractor does not scale to the full graph, this reproduction preserves the semantic--property--community design but replaces DANMF with scalable, sparse, low-rank community factors computed from the same graph.

\subsection{Main Results}

\begin{table*}[!t]
\centering
\scriptsize
\caption{Final-model comparison across the three evaluation scopes. TwiBot-22 uses the local comparison setting, TwiBot-20 uses MGTAB-style protocol-matched runs, and MGStBot-large uses three-seed external evaluation with protocol-matched non-graph runs where available. Results are reported as mean $\pm$ standard deviation. Best values among the reported rows within each dataset are bolded; ``--'' indicates that the method was not evaluated under that scope.}
\label{tab:main-results}
\setlength{\tabcolsep}{2.2pt}
\renewcommand{\arraystretch}{1.08}
\resizebox{0.98\textwidth}{!}{%
\begin{tabular}{l|ccc|ccc|ccc}
\hline
Method 
& \multicolumn{3}{c|}{TwiBot-22} 
& \multicolumn{3}{c|}{TwiBot-20} 
& \multicolumn{3}{c}{MGStBot-large} \\
\cline{2-10}
& ACC & F1 & AUC 
& ACC & F1 & AUC 
& ACC & F1 & AUC \\
\hline
Kantepe-LR 
& $0.5389{\pm}0.0453$ & $0.4765{\pm}0.0519$ & $0.5752{\pm}0.0625$ 
& $0.8183{\pm}0.0032$ & $0.8560{\pm}0.0021$ & $0.8204{\pm}0.0012$ 
& $0.8863{\pm}0.0072$ & $0.8034{\pm}0.0102$ & $0.9417{\pm}0.0013$ \\

LightGBM 
& $0.6334{\pm}0.0374$ & $0.5665{\pm}0.0015$ & $0.7781{\pm}0.0421$ 
& $0.8115{\pm}0.0047$ & $0.8504{\pm}0.0015$ & $0.8298{\pm}0.0009$ 
& $0.9098{\pm}0.0054$ & $0.8497{\pm}0.0093$ & $\mathbf{0.9681{\pm}0.0015}$ \\

RoBERTa-direct 
& $0.6993{\pm}0.0018$ & $0.5477{\pm}0.0024$ & $0.7465{\pm}0.0312$ 
& $0.7143{\pm}0.0096$ & $0.7652{\pm}0.0033$ & $0.7890{\pm}0.0028$ 
& $0.8258{\pm}0.0028$ & $0.7193{\pm}0.0053$ & $0.8993{\pm}0.0009$ \\

BotSCL 
& $0.6364{\pm}0.0272$ & $0.5216{\pm}0.0098$ & $0.7007{\pm}0.0217$ 
& -- & -- & -- 
& -- & -- & -- \\

HGT-strict 
& $0.5653{\pm}0.0059$ & $0.4849{\pm}0.0156$ & $0.6560{\pm}0.0012$ 
& -- & -- & -- 
& -- & -- & -- \\
\hline
BotCF-style 
& $0.5999{\pm}0.0258$ & $0.5304{\pm}0.0246$ & $0.7274{\pm}0.0098$ 
& $0.8526{\pm}0.0020$ & $0.8732{\pm}0.0023$ & $0.9253{\pm}0.0008$ 
& -- & -- & -- \\

BotRGCN 
& $0.6692{\pm}0.0025$ & $0.5771{\pm}0.0010$ & $0.7758{\pm}0.0028$ 
& $0.8608{\pm}0.0034$ & $0.8792{\pm}0.0032$ & $0.9285{\pm}0.0017$ 
& $0.8425{\pm}0.0020$ & $0.7356{\pm}0.0094$ & $0.9184{\pm}0.0012$ \\

RGT 
& $0.6666{\pm}0.0083$ & $0.5674{\pm}0.0105$ & $0.7679{\pm}0.0028$ 
& $0.8633{\pm}0.0017$ & $0.8809{\pm}0.0017$ & $0.9308{\pm}0.0012$ 
& $0.8441{\pm}0.0061$ & $0.7407{\pm}0.0052$ & $0.9183{\pm}0.0025$ \\

GCN 
& -- & -- & -- 
& $0.7946{\pm}0.0034$ & $0.8225{\pm}0.0030$ & $0.8673{\pm}0.0019$ 
& $0.7141{\pm}0.0348$ & $0.5660{\pm}0.0144$ & $0.7822{\pm}0.0006$ \\

GAT 
& -- & -- & -- 
& $0.8453{\pm}0.0030$ & $0.8643{\pm}0.0031$ & $0.9216{\pm}0.0023$ 
& $0.7716{\pm}0.0142$ & $0.5683{\pm}0.0035$ & $0.7826{\pm}0.0091$ \\

Simple-HGN 
& -- & -- & -- 
& $0.8586{\pm}0.0005$ & $0.8781{\pm}0.0023$ & $0.9302{\pm}0.0003$ 
& $0.8363{\pm}0.0034$ & $0.7339{\pm}0.0082$ & $0.9155{\pm}0.0002$ \\
\hline
\textsc{CSC} (final) 
& $\mathbf{0.7248{\pm}0.0065}$ & $\mathbf{0.5956{\pm}0.0023}$ & $\mathbf{0.7809{\pm}0.0014}$ 
& $\mathbf{0.8673{\pm}0.0022}$ & $\mathbf{0.8828{\pm}0.0025}$ & $\mathbf{0.9334{\pm}0.0008}$ 
& $\mathbf{0.9157{\pm}0.0051}$ & $\mathbf{0.8532{\pm}0.0107}$ & $0.9621{\pm}0.0006$ \\
\hline
\end{tabular}
}
\end{table*}

Table~\ref{tab:main-results} summarizes the final-model comparison across the three evaluation scopes. 
On TwiBot-22, \textsc{CSC} achieves the strongest operating-point performance among the reported rows, improving both accuracy and F1 over representative graph-based, semantic, and tabular baselines. 
This result is important because TwiBot-22 is the primary setting where semantic, structural, and attribute evidence are more likely to disagree. 
On TwiBot-20 and MGStBot-large, \textsc{CSC} also obtains the highest F1 among the reported rows, suggesting that the proposed framework remains competitive beyond the primary TwiBot-22 setting.

The AUC results are more nuanced. 
On TwiBot-22 and TwiBot-20, \textsc{CSC} obtains the highest AUC among the reported rows, but the margin is small on TwiBot-20. 
On MGStBot-large, LightGBM retains the highest AUC. 
Therefore, the main empirical gain of \textsc{CSC} should not be interpreted as universal ranking-metric dominance. 
Rather, the results support a more focused conclusion: calibrated evidence integration improves operating-point decision quality under modality conflict. 
This distinction is especially relevant for hard bot-detection benchmarks, where the final decision depends not only on how accounts are ranked, but also on how heterogeneous confidence scores are calibrated and thresholded.

The MGStBot-large results further clarify the scope of this conclusion. 
\textsc{CSC} improves the F1 operating point over the official large-graph baselines and strong non-graph baselines, while LightGBM remains stronger in AUC. 
The RoBERTa-direct entry on this benchmark should be read as a protocol-matched frozen-text counterpart rather than a new end-to-end RoBERTa fine-tuning run. 
This pattern suggests that \textsc{CSC} is most effective at improving the final calibrated decision, while some baselines may still retain stronger threshold-free ranking ability in specific external settings.

\begin{figure}[t]
\centering
\includegraphics[width=\columnwidth]{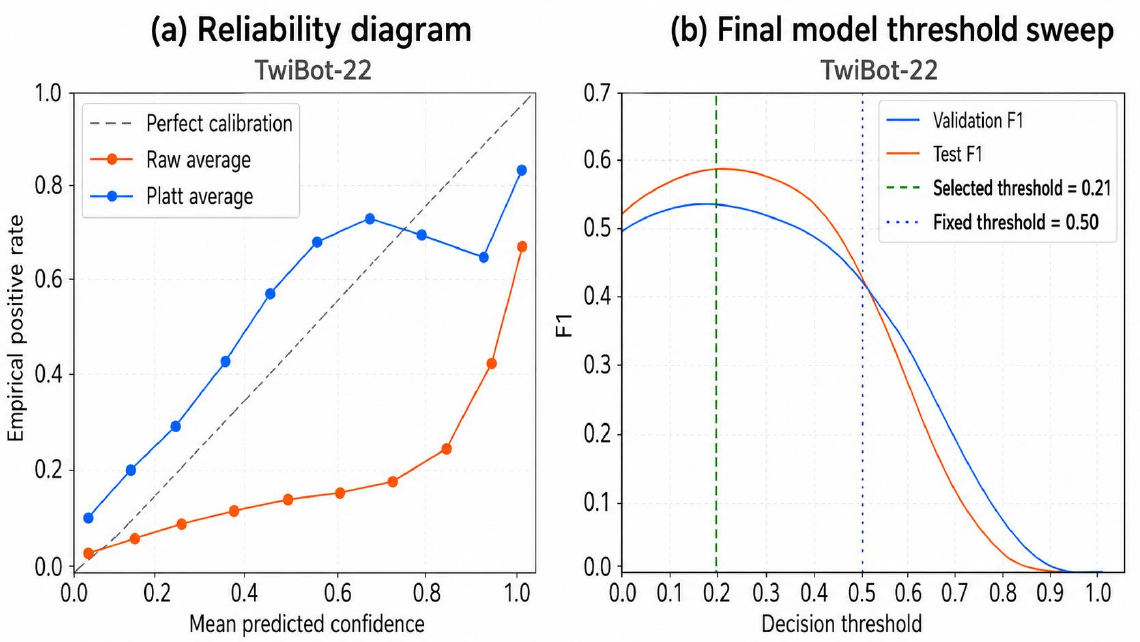}
\caption{Protocol-controlled TwiBot-22 diagnostics. 
Left: Platt calibration improves probability reliability under a fixed fusion setup. 
Right: the final fused model is threshold-sensitive and peaks near the validation-selected threshold rather than at the default threshold of $0.5$.}
\Description{Two protocol-controlled diagnostic plots for TwiBot-22. The left plot compares probability reliability before and after Platt scaling. The right plot shows F1 as a function of the decision threshold, with the final fused model peaking near the validation-selected threshold rather than the default threshold of 0.5.}
\label{fig:calibration-threshold}
\end{figure}

Figure~\ref{fig:calibration-threshold} further illustrates why calibration and threshold selection are part of the final decision protocol. 
Platt scaling improves probability reliability even when ranking changes are small, and the threshold curve shows that the final fused model peaks near the validation-selected threshold rather than at the default threshold of 0.5. 
Taken together, these results support the central claim that \textsc{CSC} improves calibrated decision quality under modality conflict, rather than simply increasing model complexity or uniformly improving threshold-free ranking metrics.

\subsection{Base Expert and Calibration Diagnostics}
\begin{figure}[t]
\centering
\includegraphics[width=\columnwidth]{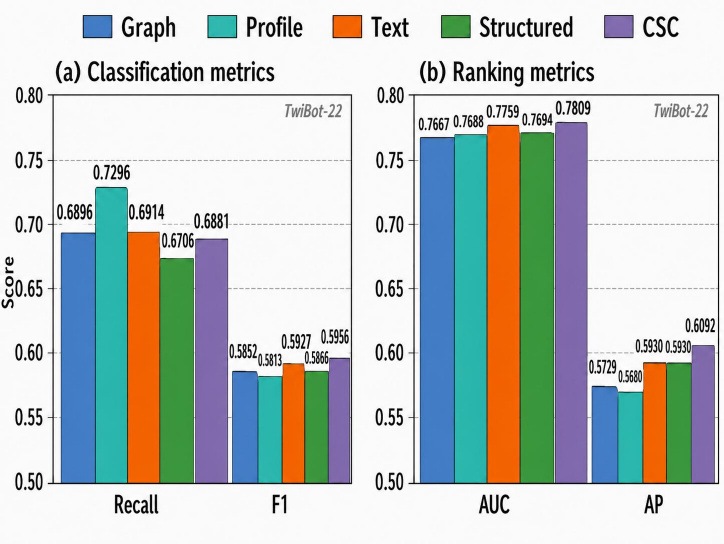}
\caption{Comparison between the four base experts and the final \textsc{CSC} model on TwiBot-22. 
The figure reports Recall and F1 for classification behavior, and AUC and AP for ranking-oriented behavior. 
Among the base experts, the profile expert achieves the highest recall. 
The final \textsc{CSC} model obtains the strongest F1, AUC, and AP, showing that calibrated fusion improves final decision quality beyond any single expert.}
\Description{A two-panel grouped bar chart comparing the calibrated graph, profile, text, and structured-feature experts with the final \textsc{CSC} model on TwiBot-22. The left panel reports Recall and F1, and the right panel reports AUC and AP.}
\label{fig:base-experts}
\end{figure}

Before analyzing the final fusion stage, we first examine whether the gain of \textsc{CSC} comes from a single dominant expert or from calibrated evidence integration. 
Figure~\ref{fig:base-experts} compares the four base experts with the final \textsc{CSC} model on TwiBot-22 under the same validation-threshold protocol. 
The base experts include the graph expert $q_i^G$, profile expert $q_i^P$, text expert $q_i^T$, and structured-feature expert $q_i^D$. 
The figure reports Recall and F1 for classification behavior, and AUC and AP for ranking-oriented behavior, where AP measures the ranking precision of the positive bot class.

Among the base experts, the profile expert obtains the highest recall, while the text expert achieves the strongest base-expert F1 and AUC and ties with the structured-feature expert on AP. 
The graph and structured-feature experts remain close to the text expert on several metrics, suggesting that non-textual evidence still carries useful signal when semantic evidence may be camouflaged. 
More importantly, the final \textsc{CSC} model achieves the strongest F1, AUC, and AP overall. 
This suggests that the final gain comes from calibrating and integrating heterogeneous evidence rather than simply selecting a single dominant expert.

These results motivate calibrated fusion rather than selecting a single best expert. 
Since no individual base expert dominates all aspects of the evaluation, the final model should combine their scores and align their confidence scales before fusion. 
Table~\ref{tab:calibration_diagnostics} further reports calibration and fusion diagnostics on TwiBot-22. 
Raw averaging and raw simplex fusion obtain similar AUC values to the calibrated variants, but their calibration errors are much higher. 
After Platt scaling, ECE, Brier score, and NLL decrease substantially, indicating that calibration mainly improves probability reliability rather than threshold-free ranking. 
The final calibrated system achieves the strongest F1 and the lowest ECE among the compared fusion variants, supporting our interpretation that \textsc{CSC} improves calibrated operating-point decision quality.
\begin{table*}[t]
\centering
\caption{Calibration and fusion diagnostics on TwiBot-22. Calibration mainly improves reliability and operating-point decision quality, while AUC changes only slightly.}
\label{tab:calibration_diagnostics}
\small
\setlength{\tabcolsep}{6pt}
\renewcommand{\arraystretch}{1.08}
\begin{tabular}{lccccccc}
\toprule
Method & F1 & AUC & AP & ECE & Brier & NLL & Thr. \\
\midrule
Raw Average & 0.5887 & 0.7807 & 0.6126 & 0.2626 & 0.2538 & 0.7333 & 0.68 \\
Raw Simplex & 0.5908 & 0.7808 & 0.6122 & 0.2643 & 0.2547 & 0.7370 & 0.70 \\
Platt Average & 0.5918 & 0.7807 & 0.6094 & 0.0899 & 0.1696 & 0.5532 & 0.18 \\
Platt Simplex & 0.5953 & \textbf{0.7809} & 0.6092 & 0.0905 & 0.1696 & 0.5530 & 0.20 \\
Platt Simplex + $C$ & \textbf{0.5956} & \textbf{0.7809} & 0.6092 & \textbf{0.0855} & \textbf{0.1687} & \textbf{0.5473} & 0.21 \\
\bottomrule
\end{tabular}
\end{table*}

\subsection{Ablation and Simplicity Analysis}

The ablation study separates graph-side architectural choices from fusion-side decision choices. 
Table~\ref{tab:graph-ablation} examines whether additional graph-branch controls consistently improve the retained graph expert, while Table~\ref{tab:system-ablation} isolates the effect of calibration, simplex fusion, and validation-selected thresholding. 
Together, these two studies clarify which components contribute to the final operating point and which forms of added complexity do not provide a consistent payoff.

\subsubsection{Graph-Branch Ablation}

\begin{table*}[!t]
\centering
\scriptsize
\caption{Graph-branch ablation under the shared validation-threshold protocol. Rows separate graph architectural choices from later fusion choices, testing whether additional graph-side controls consistently improve the retained graph expert.}
\label{tab:graph-ablation}
\resizebox{\textwidth}{!}{%
\begin{tabular}{llcccccc}
\hline
ID & Version & F1$_{20}$ & AUC$_{20}$ & Thr.$_{20}$ & F1$_{22}$ & AUC$_{22}$ & Thr.$_{22}$ \\
\hline
A1 & Text-only + GraphSAGE 
& 0.7669 & 0.8102 & 0.3300 & 0.5359 & 0.7236 & 0.3300 \\

A2 & $[x_i^{\mathrm{text}};x_i^{\mathrm{struct}}]$ + single-path GraphSAGE 
& 0.8751 & 0.9243 & 0.3400 & 0.5780 & 0.7650 & 0.6800 \\

A3 & $[x_i^{\mathrm{text}};x_i^{\mathrm{struct}}]$ + dual encoders 
& 0.8772 & 0.9270 & 0.5000 & 0.5755 & 0.7664 & 0.6600 \\

A4 & Single prototype 
& 0.8774 & 0.9241 & 0.5700 & \textbf{0.5839} & 0.7621 & 0.6700 \\

A5 & Multi-prototype, without top-$r$ 
& 0.8815 & 0.9252 & 0.5400 & 0.5822 & \textbf{0.7675} & 0.7400 \\

A6 & Multi-prototype + top-$r$=2, without residual 
& 0.8812 & 0.9251 & 0.6000 & 0.5799 & 0.7657 & 0.6200 \\

A7 & Residual + fixed weight 
& \textbf{0.8831} & 0.9259 & 0.4700 & 0.5836 & 0.7656 & 0.7200 \\

A8 & Full graph model 
& 0.8773 & \textbf{0.9267} & 0.3100 & 0.5802 & 0.7635 & 0.7000 \\
\hline
\end{tabular}
}
\end{table*}

Table~\ref{tab:graph-ablation} reports the graph-branch ablation under the shared validation-threshold protocol. 
The prototype-guided family improves over the simple GraphSAGE baseline, confirming that structural prototype evidence is useful for social bot detection. 
Adding structured features and dual encoders also improves the graph branch over the text-only GraphSAGE baseline, but the later variants show that more graph-side control logic does not necessarily translate into better TwiBot-22 performance.

In particular, sparse top-$r$ prototype selection and the full graph-side control variant introduce additional design choices, yet they do not produce a clear improvement over the simpler retained variants. 
For example, the single-prototype and residual fixed-weight variants remain competitive on TwiBot-22, while the full graph model does not achieve the strongest F1 under the shared protocol. 
This supports the narrower simplicity claim of the paper: the goal is not to make the graph branch minimal, but to preserve reliable structural biases while removing graph-side control mechanisms that do not yield stable gains.

\subsubsection{Fusion-Stage Ablation}

\begin{table*}[!t]
\centering
\scriptsize
\caption{Fusion-stage ablation under the shared local protocol. S5 is the final validation-threshold system; S6 uses the same fused scores as S5 but fixes the decision threshold at $0.5$.}
\label{tab:system-ablation}
\resizebox{\textwidth}{!}{%
\begin{tabular}{llcccccc}
\hline
ID & Version & F1$_{20}$ & AUC$_{20}$ & Thr.$_{20}$ & F1$_{22}$ & AUC$_{22}$ & Thr.$_{22}$ \\
\hline
S1 & Best graph branch 
& 0.8839 & 0.9256 & 0.5200 & 0.5802 & 0.7635 & 0.7000 \\

S2 & Best non-graph branch & 0.8820 & 0.9327 & 0.4600 & 0.5903 & 0.7759 & 0.7100 \\

S3 & Graph + non-graph, direct average 
& 0.8779 & 0.9328 & 0.5800 & 0.5907 & 0.7763 & 0.7000 \\

S4 & Graph + non-graph + Platt calibration 
& 0.8831 & 0.9321 & 0.4900 & 0.5918 & 0.7762 & 0.1700 \\

S5 & Final calibrated fusion + validation threshold 
& 0.8827 & \textbf{0.9334} & 0.4800 & \textbf{0.5940} & \textbf{0.7801} & 0.1900 \\

S6 & Final calibrated fusion + fixed threshold 0.5 
& 0.8809 & \textbf{0.9334} & 0.5000 & 0.4362 & \textbf{0.7801} & 0.5000 \\
\hline
\end{tabular}
}
\end{table*}

Table~\ref{tab:system-ablation} further isolates the fusion stage. 
The strongest non-graph branch is already competitive, which shows that the final improvement cannot be attributed simply to adding a weak auxiliary signal to a dominant graph expert. 
Direct averaging and Platt-calibrated fusion both improve over relying on the graph branch alone, but the final calibrated fusion with validation-threshold selection gives the strongest TwiBot-22 operating point in this controlled comparison.

The contrast between S5 and S6 is particularly important. 
The two rows use the same fused scores and therefore have the same AUC, but fixing the decision threshold at $0.5$ substantially reduces F1. 
This result explains why threshold selection is treated as part of the method rather than as a minor implementation detail. 
It also supports the main empirical interpretation of \textsc{CSC}: the framework mainly improves calibrated operating-point decision quality, rather than simply increasing threshold-free ranking performance.

Finally, the simplified final formulation preserves the performance reported in Table~\ref{tab:main-results}. 
Relative to the calibrated base blend without the inconsistency expert, the aggregate gain is modest. 
We therefore interpret the inconsistency expert as a lightweight disagreement-aware corrector rather than as the main source of the benchmark improvement.

\subsection{Semantic Camouflage Stress Test}
\label{sec:semantic-camouflage}

We further evaluate whether the proposed framework is robust to the semantic-camouflage failure mode that motivates this paper. 
In the TwiBot-22 stress test, selected bot accounts keep their graph structure and non-text attributes unchanged, while their textual evidence is replaced with matched human text. 
This intervention weakens semantic evidence without changing the surrounding relational, profile, or structured-feature evidence.

On the balanced semantic-camouflage challenge set, the text branch is highly sensitive to this perturbation.
After matched-human text replacement, its F1 drops from 0.9217 to 0.6780, and its recall drops from 1.0000 to 0.6000. 
In contrast, the graph branch and the fused systems remain stable on the balanced challenge set. 
This result does not imply that graph evidence is always superior to textual evidence. 
Rather, it shows that text can be a strong but fragile signal, and that non-textual evidence can provide an important anchor when semantic cues become easier to imitate.

The comparison between base fusion and the final system also clarifies the scope of the inconsistency expert. 
In this stress test, the final system does not obtain a large additional gain over base fusion. 
Therefore, the stress test mainly supports the multimodal robustness argument: when bot text is deliberately made more human-like, calibrated fusion of semantic and non-semantic evidence is more stable than relying on the text branch alone. 
The inconsistency expert should instead be interpreted as a lightweight correction module for disagreement-heavy regions, which we examine in Section~\ref{sec:conflict-region}.

To avoid relying only on the local semantic-camouflage proxy, we also conduct a reduced BotSim-24 sanity check~\cite{qiao2025botsim} on public LLM-driven bot data. 
We exclude user identifiers, row order, names, and bot character settings, and use only profile/activity text, public behavior statistics, and a lightweight comment-interaction graph. 
Across three stratified splits, the text branch reaches 0.9942 $\pm$ 0.0057 F1, the behavior/graph feature branch reaches 0.9868 $\pm$ 0.0075 F1, and conflict-aware fusion reaches 0.9983 $\pm$ 0.0029 F1. 
Since text alone is already very strong on this Reddit-based benchmark, we interpret this result as auxiliary LLM-era sanity evidence rather than as the main evidence for the inconsistency expert.
\subsection{Conflict-Region Analysis}
\label{sec:conflict-region}

We further analyze where the inconsistency expert contributes. 
As shown in the ablation results, its aggregate gain over the calibrated base fusion is modest. 
Therefore, we do not interpret it as a strong standalone detector or as the main source of the final benchmark improvement. 
Instead, we examine whether it provides localized corrections in disagreement-heavy regions.

We compute a disagreement score from the variance, range, and graph-centered gaps among calibrated expert scores, and divide test samples into low-, medium-, and high-conflict groups. 
Table~\ref{tab:conflict-region} shows that the correction effect is concentrated in the high-conflict group. 
Although the F1 gain is small in aggregate, the number of corrected decisions is larger than the number of newly introduced errors in the high-conflict region. 
This pattern supports the role of the inconsistency expert as a lightweight disagreement-aware corrector. We also examine near-threshold samples, where small changes in calibrated evidence are more likely to affect the final decision. 
The same pattern appears: the inconsistency expert corrects more near-boundary cases than it hurts. 
These results suggest that cross-modal disagreement is most useful as localized correction evidence, rather than as a replacement for calibrated fusion.
\begin{table}[t]
\centering
\caption{Conflict-region analysis on TwiBot-22. Samples are split into low-, medium-, and high-conflict terciles by disagreement score; $N$ denotes the number of test samples in each region. The inconsistency expert mainly helps in the high-conflict region.}
\label{tab:conflict-region}
\small
\resizebox{\columnwidth}{!}{%
\begin{tabular}{lrrrrrr}
\toprule
Region & $N$ & Base F1 & +$C$ F1 & $\Delta$F1 & Fix & Hurt \\
\midrule
Low conflict    & 33,333 & 0.3687 & 0.3687 & -0.0001 & 2  & 3  \\
Medium conflict & 33,333 & 0.6478 & 0.6478 & -0.0000 & 4  & 9  \\
High conflict   & 33,334 & 0.5975 & 0.5981 &  0.0006 & 96 & 38 \\
\bottomrule
\end{tabular}%
}
\end{table}
\section{Conclusion}

We presented \textsc{CSC}, a conflict-aware social bot detection framework for the LLM era. 
The method combines a simplified prototype-guided graph expert, validation-only calibrated simplex fusion, and a lightweight inconsistency expert. 
Instead of continually adding graph-side control logic, \textsc{CSC} preserves reliable structural biases, aligns heterogeneous confidence scores, and uses cross-modal disagreement as localized correction evidence.
Experiments across TwiBot-22, TwiBot-20, and MGStBot-large show that \textsc{CSC} improves calibrated operating-point decision quality while remaining competitive across external benchmarks. 
The calibration diagnostics show that probability, reliability, and threshold selection are important for hard bot-detection settings. 
The semantic-camouflage stress test further shows that textual evidence can be strong but fragile when bot text is made more human-like, while graph and fused evidence remain more stable. 
Finally, the conflict-region analysis indicates that the inconsistency expert mainly acts as a lightweight corrector in disagreement-heavy or near-threshold cases rather than as the primary source of aggregate performance gain.
Overall, our results suggest that robust bot detection in the LLM era should treat cross-modal conflict as a central decision problem. 
Under this setting, preserving useful structural signals and calibrating heterogeneous evidence can be more reliable than simply increasing graph-side complexity.

\section{Limitations and Future Work}

Several limitations should be considered when interpreting our results. 
First, although we organize the evaluation by protocol scope and provide protocol-controlled diagnostics, not all external baselines are rerun under a single unified codebase. 
We therefore treat broad benchmark comparisons as contextual and rely on controlled local diagnostics for mechanism-level conclusions. 
Future work should evaluate \textsc{CSC} under a fully unified pipeline across benchmarks such as TwiBot-20, TwiBot-22, and MGTAB~\cite{feng2021twibot20,feng2022twibot22,shi2023mgtab}.

Second, the final detector is sensitive to the operating point on hard benchmarks such as TwiBot-22. 
Accordingly, validation-selected thresholding is treated as part of the decision protocol, and we report calibration diagnostics beyond threshold-free metrics. 
More robust threshold transfer across datasets, time periods, and deployment settings remains an important direction.

Third, the semantic-camouflage stress test isolates text-side camouflage and should be viewed as a controlled proxy rather than a complete simulation of realistic LLM-driven bots. 
Although BotSim-24 provides auxiliary evidence~\cite{qiao2025botsim}, larger platform-specific evaluations are still needed. 
Cross-modal disagreement may also arise from missing information, modality-specific noise, or unequal expert quality rather than bot behavior itself. 
Moreover, static graph prototypes may not capture evolving structural camouflage, and our current stress test does not consider simultaneous manipulation of structural or profile evidence.

Finally, the lightweight inconsistency expert may not capture richer temporal, conversational, or group-level disagreement. 
High-concurrency inference latency also remains to be systematically evaluated. 
Future work will investigate adaptive modality reliability, temporally updated structural modeling, richer disagreement signals, and deployment efficiency.

\section*{GenAI Usage Disclosure}
Generative AI assistance was used for editorial support during manuscript preparation, including wording refinement and formatting iteration. It was not used to generate experimental results or labels. All technical claims, reported numbers, and final manuscript decisions were verified by the authors.

\bibliographystyle{ACM-Reference-Format}
\bibliography{refs}

\end{document}